\documentclass{article}

\usepackage{PRIMEarxiv}
\usepackage{eccvabbrv}
\usepackage{multirow}
\usepackage[utf8]{inputenc} 
\usepackage[T1]{fontenc}    
\usepackage{hyperref}       
\usepackage{url}            
\usepackage{booktabs}       
\usepackage{amsfonts}       
\usepackage{amsmath}
\usepackage{amssymb}
\usepackage{amsthm}
\usepackage{nicefrac}       
\usepackage{microtype}      
\usepackage{lipsum}
\usepackage{fancyhdr}       
\usepackage{graphicx}       
\usepackage{color}
\usepackage{xcolor}
\usepackage{colortbl}
\usepackage{subcaption}
\usepackage{wrapfig}
\usepackage[capitalize]{cleveref}
\graphicspath{{media/}}     
\newcommand{\ours}{\textsc{VLN-GPT}}
\title{Vision-and-Language Navigation Generative Pretrained Transformer}

\author{
  Hanlin Wen \\
  School of Artificial Intelligence and Automation \\
  Huazhong University of Science and Technology \\
  Wuhan\\
  \texttt{Hennywhen@gmail.com} 
}

\begin{document}
\maketitle

\begin{abstract}
In the Vision-and-Language Navigation (VLN) field, agents are tasked with navigating real-world scenes guided by linguistic instructions. Enabling the agent to adhere to instructions throughout the process of navigation represents a significant challenge within the domain of VLN. To address this challenge, common approaches often rely on encoders to explicitly record past locations and actions, increasing model complexity and resource consumption.
  Our proposal, the Vision-and-Language Navigation Generative Pretrained Transformer (VLN-GPT), adopts a transformer decoder model (GPT2) to model trajectory sequence dependencies, bypassing the need for historical encoding modules. This method allows for direct historical information access through trajectory sequence, enhancing efficiency. Furthermore, our model separates the training process into offline pre-training with imitation learning and online fine-tuning with reinforcement learning. This distinction allows for more focused training objectives and improved performance.  
  Performance assessments on the VLN dataset reveal that VLN-GPT surpasses complex state-of-the-art encoder-based models.
\end{abstract}
\begin{wrapfigure}{r}{0.4\textwidth} 
  \centering
  \vspace{-0.8cm}
  \includegraphics[width=0.4\textwidth]{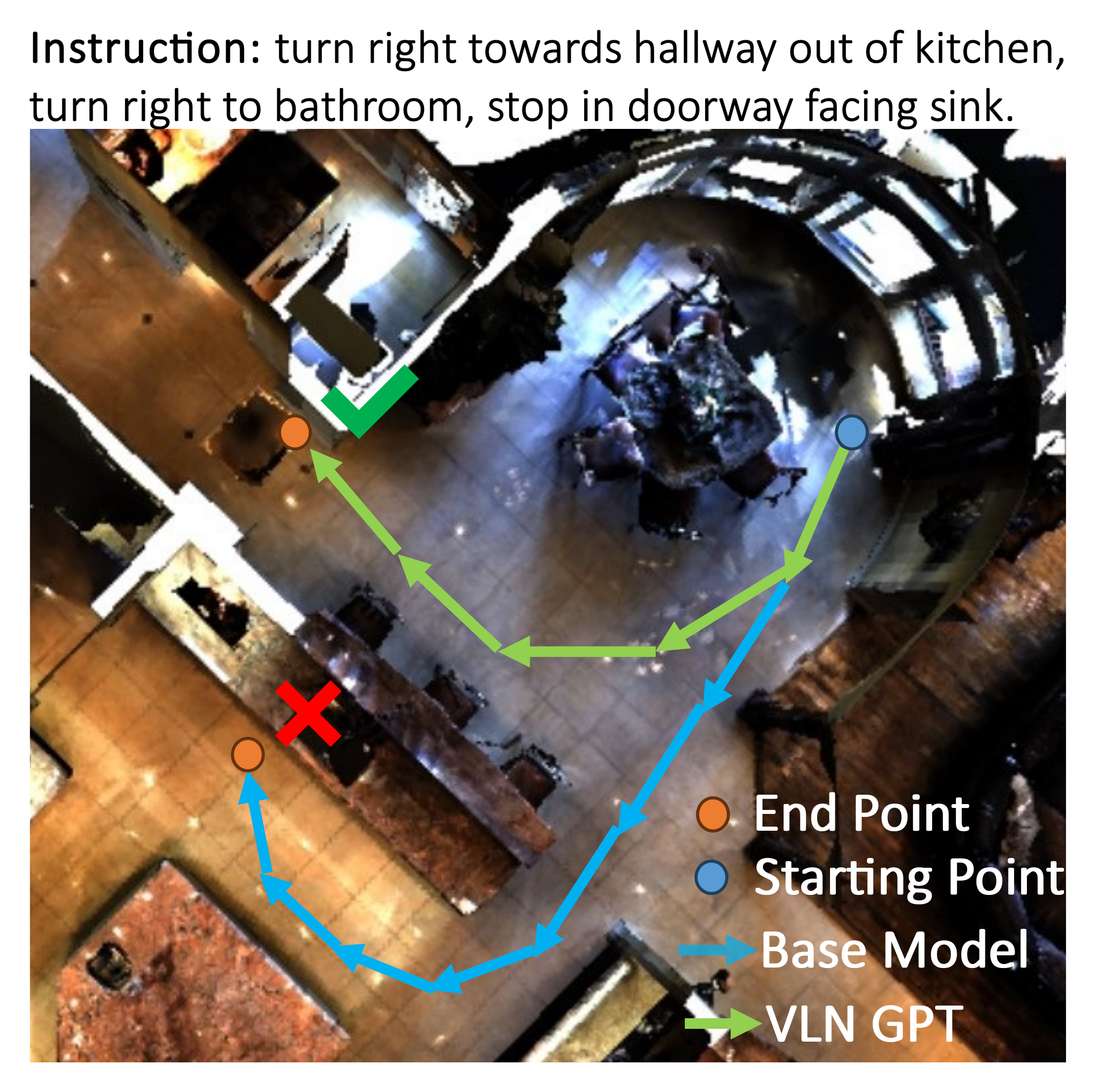}
   \caption{ Demonstration of an example from the R2R validation dataset \cite{r2r}.
   More details of qualitative results seen in the Appendix.
  }
  \vspace{-0.16cm}
\end{wrapfigure}

\section{Introduction}
The advent of large language models\cite{llama,gpt2,gpt3model} and multi-modal models\cite{li2022blip,li2023blip2} represents a significant stride towards the realization of artificial general intelligence (AGI)\cite{goertzel2014artificial}. Among the diverse pathways toward AGI, Vision-and-Language Navigation (VLN)\cite{anderson2018evaluation} stands out as a critical area of focus within the Embodied Agent community\cite{embodied}. This domain necessitates agents to navigate adeptly in photo-realistic environments guided by natural language instructions.

A paramount challenge in Vision-and-Language Navigation (VLN) involves the retention of sequential observations and feedback. Different from other Vision-Language tasks like Vision Question Answering (VQA)\cite{VQA}, where the imagery remains static, VLN entails a dynamic visual context that evolves over time through the navigation process. Early attempts\cite{zhu2020babywalk,r2r,rnnbert} leveraged Recurrent Neural Networks (RNNs)\cite{rnn} to encapsulate these changing environments, namely encoding observations and actions within a compact state vector to facilitate subsequent action prediction. However, the inherent limitations of RNNs—particularly their tendency to overlook initial states in longer trajectories—restrict their applicability for the nuanced navigation sequences in VLN. In response, subsequent studies\cite{graphmem,structuremem} introduced memory modules, employing a map-like mechanism to archive sequential observations, though still relying on RNNs for state tracking. With the transformative success of the transformer architecture\cite{transformer}, recent research has explored its integration into VLN. Transformers, with their adeptness at capturing long-term sequence dependencies, encode historical data as sequences of past actions and observations. While the transformer encoder addresses RNNs' drawbacks, it also introduces increased model complexity and computational demands.

Training paradigms present another significant challenge in Vision-and-Language Navigation (VLN). Reinforcement Learning (RL) \cite{reinforcement} is widely adopted to refine navigation policies, integrating techniques such as Imitation Learning (IL) and Asynchronous Advantage Actor-Critic (A3C) in several studies\cite{tan-etal-2019-learning,hamt,gela}. Although these methods have shown efficacy, achieving a balance between exploration and exploitation remains a daunting task in RL. Specifically, IL aims to guide the agent in emulating expert behaviors, while RL motivates exploration based on the learning policy, leading to an intrinsic conflict between these objectives. Thus, devising a strategy that effectively merges these divergent goals is crucial. Current approaches tend to blend these elements with static hyperparameters throughout training, which is suboptimal.

Mirroring the success observed in natural language processing, the pre-train and fine-tune paradigm has been adopted in VLN research\cite{PREVALENT,hamt,gela}. The initial goal of pre-training is to cultivate a robust representation of visual and linguistic inputs. To this end, a variety of pre-training proxy tasks and losses, such as Single-step Action Prediction and Spatial Relationship Prediction \cite{PREVALENT,hamt}, are employed. These elements undoubtedly add to the training complexity, making the amalgamation of multiple pre-training losses as challenging as integrating IL and RL.

\begin{figure}[!t]
    \centering
    \includegraphics[width=\textwidth]{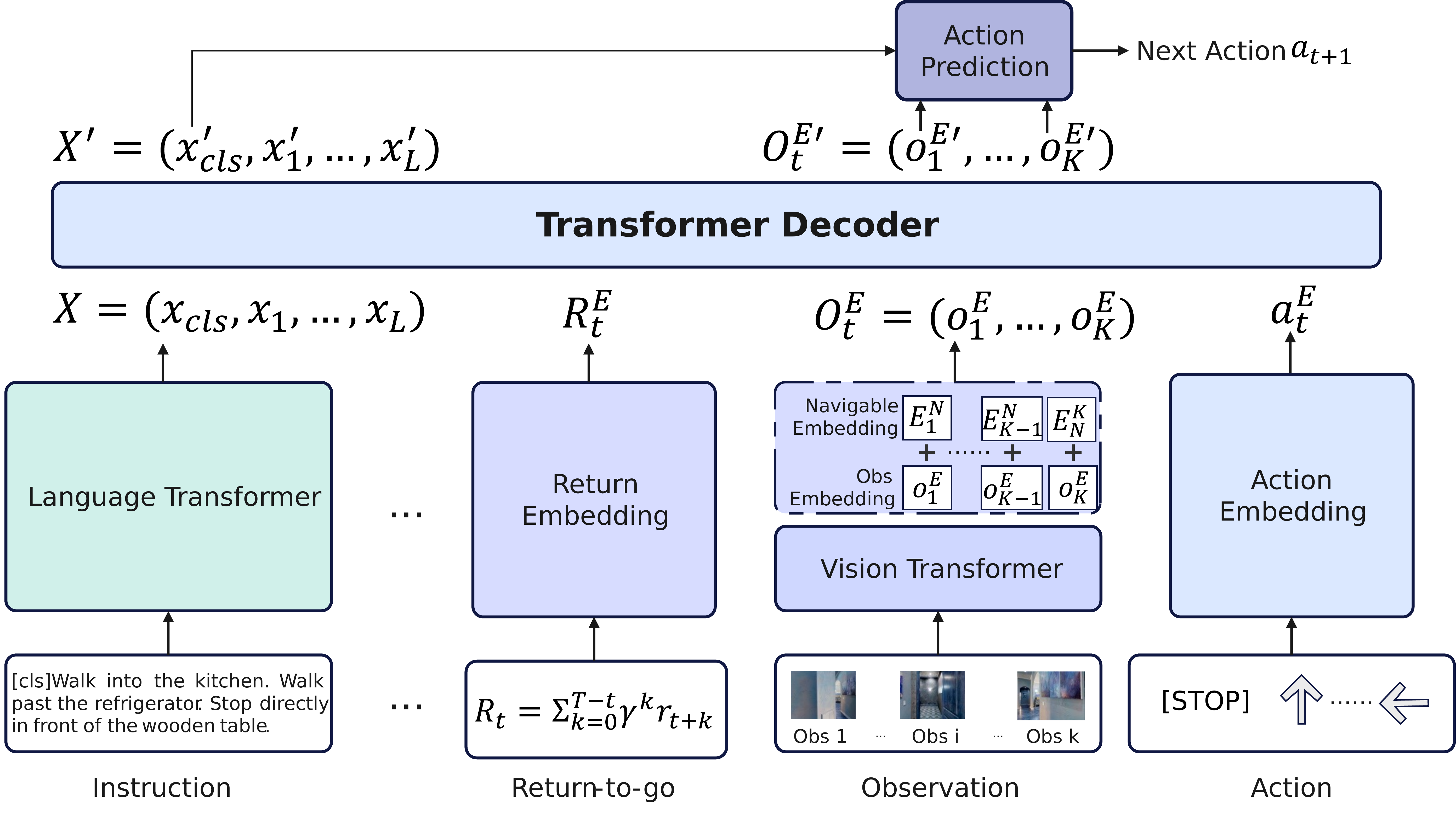} 
    \caption{
    The architecture of \textbf{V}ision-and-\textbf{L}anguage \textbf{N}avigation \textbf{G}enerative \textbf{P}retrained \textbf{T}ransformer, namely \ours. 
    \ours\ adopts a transformer decoder to model the dependencies of instruction, returns, observations, and actions in the trajectory sequence and predict action on the observation token at each time step $t$.
    }
    \label{fig:architecture}
\end{figure}

To overcome the challenges outlined previously, we propose the Vision and Language Navigation Generative Pretrained Transformer (VLN-GPT) model, a decoder-only transformer architecture for multi-modal decision-making in the VLN task.
As depicted in Figure \ref{fig:architecture}, VLN-GPT incorporates a BERT-based text embedding module \cite{sentencebert}, a Vision Transformer (ViT)-based observation embedding module \cite{vit}, and a GPT-2-based transformer decoder architecture \cite{gpt2} to delineate the dependencies between instructions and observations in the trajectory sequence. Owing to the inherent capacity of this architecture to integrate the history of each observation within the sequence, explicit encoding of the historical sequence is rendered unnecessary, thereby economizing on computational resources. Additionally, this approach streamlines the model by leveraging the GPT model's robust sequence processing prowess. Through the masked attention mechanism \cite{transformer}, VLN-GPT is restricted to only reference preceding observations and actions, mimicking the historical information encoder commonly adopted in transformer-encoder-based methods.
In addressing the complexity inherent in the training process, we delineate the objectives of exploration and exploitation during the fine-tuning stages. Specifically, we streamline the training objectives of proxy tasks and the corresponding losses in the pertaining phase. We adopt offline reinforcement learning to supervise the model based on expert trajectories during pre-training, focusing exclusively on the single-step action prediction task. This strategy renders the pre-training phase both more focused and efficient.
Unlike transformer-encoder-based methods that are limited to learning representations during the pretraining stage, our model is capable of further understanding the multi-modal dependencies between instructions and trajectories within this phase.
Moreover, to foster exploration during the online fine-tuning stage, we utilize the entropy of the policy as inspired by Zheng et al.\cite{zheng2022online}, enhancing the model's ability to navigate in novel environments.

Empirically, our evaluation involves conducting experiments on the Room-to-Room (R2R) dataset and benchmarking our method against state-of-the-art algorithms to substantiate its efficacy. Our findings indicate that our approach outperforms the more complex and computationally demanding transformer-encoder-based methods.

Our contributions are summarized as follows: (1) We pioneer the sequential modeling approach for Vision-and-Language Navigation (VLN) tasks, and introduce VLN-GPT, a decoder-only transformer architecture specifically designed for multi-modal decision-making within this domain. (2) We innovatively segregate the objectives of exploration and exploitation, allocating them to offline pre-training and online fine-tuning stages, respectively. (3) Our methodology is rigorously validated against the state-of-the-art (SOTA) transformer-encoder-based approaches, demonstrating promising performance outcomes. 
\section{Related Work}
\textbf{Vision-and-language navigation.} Since Anderson \etal introduced the Room-to-Room (R2R) dataset\cite{r2r}, Vision-and-Language Navigation (VLN) has garnered considerable interest within the academic community. Alongside the R2R dataset, an LSTM-based benchmark model for the VLN task was concurrently developed. Subsequent to this foundational work, a series of studies employing LSTM\cite{lstm} architecture models emerged\cite{zhu2020babywalk,fried2018speaker,ma2019selfmonitoring}, further advancing the field. Furthermore, Reinforcement Learning (RL) is commonly utilized to refine navigation policies, with numerous distinguished VLN models embracing both imitation and reinforcement learning-based training paradigms. Notably, Babywalk\cite{zhu2020babywalk} integrates imitation learning with curriculum learning to train the agent effectively.
The EnvDrop\cite{tan-etal-2019-learning} model amalgamates IL with A3C \cite{mnih2016asynchronous}. Given the transformer model's monumental success in natural language processing, recent endeavors have sought to integrate this architecture into Vision-and-Language Navigation (VLN) tasks. This has led to a proliferation of studies proposing transformer-based approaches.
PRESS\cite{li-etal-2019-robust} innovates by substituting the LSTM-based instruction encoder with a pre-trained BERT model\cite{devlin2019bert}, marking a significant shift towards leveraging advanced language understanding capabilities. SIA\cite{Lin_2021_CVPR} integrates a transformer for single-step multi-modal fusion, although it retains the LSTM architecture for action prediction, blending traditional and modern approaches. PTA\cite{landi2021multimodal} employs a transformer-based model for action prediction at each timestep, yet it continues to extract visual features via a Convolutional Neural Network (CNN), showcasing a hybrid approach to feature processing. Notably, HAMT\cite{hamt} represents the first fully transformer-based architecture for VLN, trained in an end-to-end manner, setting a new standard for architectural coherence in the domain.
While the aforementioned studies have incorporated transformer models, their utilization has predominantly been confined to extracting textual or visual features, or to the fusion of these two modalities. The transformative potential of the transformer's sequence modeling capabilities remains largely untapped.
\newline\noindent
\textbf{Historical information.} Despite the predominant reliance on Markov Decision Processes (MDP) in most studies, they still incorporate historical information. Specifically, the LSTM model is capable of encoding memories or historical records, allowing the past trajectory to be inherently included within the model for those studies employing the LSTM approach\cite{zhu2020babywalk,fried2018speaker,ma2019selfmonitoring}.
Other research initiatives have proposed alternative methodologies that integrate topological map memory structures. Deng et al.\cite{graphmem} employ graph representations to depict the environment's layout, thereby aiding in long-term strategic planning. In a similar vein, Wang et al.\cite{structuremem} embrace a graph-based strategy to assimilate frontier exploration into their decision-making process.
However, LSTM models continue to be utilized for state tracking in these studies. Yet, in light of the transformer architecture's demonstrated capability to leverage long-term temporal dependencies within sequences, Fang et al.\cite{transformermem} have implemented a transformer encoder to encode historical data meticulously. 
Furthermore, the introduction of Recurrent VLN-BERT\cite{rnnbert} represents an innovative adaptation, incorporating a transformer encoder augmented with a recurrent unit specifically for encoding historical data in the VLN task. Subsequently, Chen et al.\cite{hamt} advanced a hierarchical encoding framework for historical information, seamlessly integrating these encodings with the states for comprehensive end-to-end training.
All the aforementioned approaches utilize a dedicated module, be it an LSTM or a Transformer Encoder, to manage historical data, which inevitably escalates both the architectural and computational complexity. 
\newline\noindent
\textbf{Multi-modal pre-trained transformers.} Pretrained Transformer models such as BERT, BLIP\cite{li2022blip}, and GPT\cite{gpt3model} have garnered significant acclaim in the fields of natural language processing and computer vision, showcasing remarkable achievements. In the context of Vision-and-Language Navigation (VLN) tasks, several studies have ventured beyond the conventional use of Convolutional Neural Networks (CNNs) for image representation extraction, exploring the integration of multimodal pre-trained transformers instead.
ViLT\cite{vilt} innovatively substitutes the traditional Convolutional Neural Network (CNN) with a Vision Transformer (ViT) for visual feature extraction, facilitating training alongside associated instruction texts in an end-to-end fashion. Additionally, multiple studies have explored multi-modal pre-training approaches for Vision-and-Language Navigation (VLN). Notably, PREVALENT\cite{PREVALENT} undertakes pre-training of a transformer model using instructions and single-step observations as inputs. However, it is important to note that these efforts did not incorporate historical trajectories in the pre-training phase. 
HAMT\cite{hamt} introduced a transformer framework adept at concurrently encoding text, history, and observation. While this pre-training methodology is both potent and applicable, it is also notably time-intensive, necessitating an extensive dataset and a suite of meticulously crafted pre-training tasks. Moreover, this pre-training approach lacks flexibility, rendering the model challenging to adjust for novel tasks.
\section{\ours}
In this section, we delineate our Vision-and-Language Navigation Generative Pretrained Transformer (VLN-GPT) methodology. Initially, we present preliminary knowledge pertinent to Vision-and-Language Navigation (VLN) tasks. Subsequently, we elucidate our conceptualization of sequential modeling within the context of the VLN task. Thereafter, the architecture of the VLN-GPT model is expounded. Conclusively, we detail the dual-phase approach encompassing offline pre-training and online fine-tuning for the model.
\subsection{Preliminary}
In Vision-and-Language Navigation (VLN) tasks, an agent receives a sequence of instructions in natural language, represented as $X$, comprising multiple sentences. The agent's objective, guided by instruction $X$, involves executing a sequence of navigation actions to arrive at the designated destination. This sequence of actions, referred to as the trajectory, is denoted by $Y$. The trajectory $Y$ is composed of a sequence of state-action pairs, which is defined as:
\begin{equation}
    Y=\left(s_0, a_0,  s_1, a_1, \ldots, s_{|Y|}, a_{|Y|}\right)\,,
\end{equation}
where $|\cdot|$ denotes the sequence length or a set's size. Hence, the VLN task is to learn a mapping from the instruction $X$ to the trajectory $Y$, which is formulated as:
\begin{equation}
    X \rightarrow Y \,.
\end{equation}
In the VLN setting, the agent receives a new set of panoramic visual observation $O_t = \{o_{t,i}\}_{i=1}^{36}$ for the environment at viewpoint $O_t$ in timestep $t$, each $o_{t,i} \in O_t$ consists of an RGB image of the ith view. $V_{O_t}$ is the list of navigatable points at viewpoint $O_t$.
The action space at time $t$ is defined as $A_t = \{[STOP],a_{t,1},\ldots,a_{t,36}\}$, which consists of a set of navigational actions. The agent's action involves selecting a specific waypoint from the set $V_{O_t}$ to navigate towards. Notably, the initial action in the sequence $A_t$ is defined as the stop action $a_{t,0} = [STOP]$, signifying that the agent ceases movement at its current location, thereby concluding the navigation process. At each timestep, $t$, the agent is required to choose an action $a_t \in A_t$ that leads to the next viewpoint. Subsequently, the agent is awarded a reward $r_t$ and encounters a new observation $O_{t+1}$.
\subsection{Sequential Modelling}
\label{sec:sequential_modelling}
Contrary to the common approach in research that treats this task as a Markov decision process (MDP), our study draws inspiration from the Decision Transformer \cite{chen2021decision} and employs a sequential decision-making framework to conceptualize the problem.
The trajectory $\tau$ in the sequential decision-making process is defined by the following equation:
\begin{equation}
    \label{traject_equation}
    \tau_T=\left(r_0,s_0, a_0, r_1, s_1, a_1, \ldots,r_T, s_T, a_T \right) \,,
\end{equation}
where $T$ denotes the trajectory length and $s_t$ is the state at time step $t$, $a_t$ is the action taken at time step $t$, and $r_t$ is the return received at time step $t$.

In MDP, the probability $P$ of taking action $a_t$ when in state $s_t$ at time step $t$, is determined by the policy $\pi$, which is defined as:
\begin{equation}
    \pi(a_t|s_t)=P(a_t|s_t) \,.  
\end{equation}
While in sequnetial setting, the probability $P$ of taking action $a_t$ at time step $t$ is based on the state $s_t$ and the history of the trajectory $\tau_{t-1}$, which is defined as:
\begin{equation}
    \label{eq:sequential_policy}
    \pi(a_t|s_t,r_t, \tau_{t-1})=P(a_t\mid r_0,s_0, a_0, r_1, s_1, a_1, \ldots,r_{t-1}, s_{t-1}, a_{t-1},r_{t},s_{t})\,.
\end{equation}

Building upon the aforementioned definition, we can formulate the Vision-and-Language Navigation (VLN) task sequentially. Given an instruction $X$, an initial state $s_0$, and an average expected return $r_0$, the probability of executing a sequence of actions $A = [a_0, \ldots, a_T]$ can be determined by the product of the conditional probabilities.
\begin{equation}
    P(A)=\prod_{i=1}^n P\left(a_n \mid r_0,s_0, a_0, \ldots, r_{T-1}, s_{T-1}, a_{T-1},r_{T},s_{T}\right)\,.
\end{equation}

\subsection{Architecture}
In this section, the architecture of our Vision-and-Language Navigation Generative Pre-trained Transformer (VLN-GPT) model is discussed. Initially, the methodology for encoding input data and integrating information from diverse modalities is introduced. Subsequently, an in-depth examination of the VLN-GPT model's architecture is presented.

\noindent
\textbf{Input encoding.} Similar to studies utilizing transformer models \cite{hamt,gela}, the input instruction $X$ and the observation $O_t$ are encoded separately through their respective unimodal transformers before integration into the transformer model, which facilitates the analysis of the relationship between the instructions and the trajectories. 
Distinct from these studies, which incorporate a history encoder for the information preceding $O_t$—spanning from $O_0$ to $O_{t-1}$—our approach omits the history encoder. This modification is based on the premise that the Vision-and-Language Navigation (VLN) task, defined as a sequential decision-making process in \cref{sec:sequential_modelling}, intrinsically captures historical data within the sequence, rendering the history encoder superfluous.

\noindent
\textbf{Text encoding.} We embed the instruction $X$ through a pre-trained sentence-bert model~\cite{sentencebert}, and then obtain the contextual representation of the instruction by the [CLS] token $\mathbf{x}_{cls}$.

\noindent
\textbf{Observation encoding.} For each view $o_{t,i}$ in the observation $O_t$, we use a pre-trained vision transformer(ViT) \cite{vit}to encode the observation and then acquire the embedding of each view $o_{t,i}$, denoted as $\mathbf{o}_{t,i}^{E}$.

\noindent
\textbf{Modalities fusion.} We have devised a straightforward yet remarkably effective method for merging instruction and observation. In contrast to other studies employing a cross-modal transformer for the integration of instruction text and observation image, our approach utilizes an element-wise multiplication strategy to amalgamate the two modalities. This choice is predicated on the understanding that the embedding vectors derived from either BERT or Vision Transformer (ViT) constitute advanced representations of their respective inputs. Both embeddings share a dimensionality of 768 and possess the intrinsic ability to delineate the relationship between the instruction and the observation.
Then we can obtain the fused representation of the instruction $\mathbf{x}_{cls} \in \mathbb{R}^{1 \times 768}$ and the observation $\mathbf{o}_{t,i}^{E} \in \mathbb{R}^{1 \times 768} $, namely the state $\mathbf{s}_{t,i} \in \mathbb{R}^{1 \times 768}$, by the following equation:
\begin{equation}
    \label{eq:state}
    \mathbf{s}_{t,i} = \mathbf{x}_{cls} \odot \mathbf{o}_{t,i}^{E}\,.
\end{equation}

\noindent
\textbf{VLN GPT.} Given the formulation of the Vision-and-Language Navigation (VLN) task as a sequential decision-making process (as outlined in \cref{sec:sequential_modelling}), we harness the robust sequence processing capabilities of transformer decoder in large language models to address this problem. Following the precedent set by the Decision Transformer \cite{chen2021decision}, we adopt the GPT-2 \cite{gpt2} transformer architecture as our foundational model. To tailor it to the VLN task, we have implemented several modifications to the model.

Given a trajectory $\tau = \left(r_0, s_0, a_0, r_1, s_1, a_1, \ldots, r_T, s_T, a_T \right)$, an embedding layer is first employed to project the return, state, and action onto a unified dimensional space. Instead of utilizing the traditional positional encoding found in the standard transformer architecture, this approach incorporates a time step embedding, denoted as $v_t^p$, added to each return, state, and action embedding vector—$v_t^r$, $v_t^s$, $v_t^a$ respectively. These vectors are then concatenated to construct the input sequence.
\begin{equation}
    \begin{array}{l}
        \mathbf{v}_t^r = \text{Embedding\_Return}(r_t)\,,\\
        \mathbf{v}_t^s = \text{Embedding\_State}(s_t)\,,\\
        \mathbf{v}_t^a = \text{Embedding\_Action}(a_t)\,,\\
        \mathbf{v}_t^p = \text{Embedding\_Time\_Step}(t)\,,\\
        \mathbf{v}_t = [\mathbf{v}_t^r,\mathbf{v}_t^s,\mathbf{v}_t^a] + [\mathbf{v}_t^p,\mathbf{v}_t^p,\mathbf{v}_t^p]\,.
    \end{array}
\end{equation}

After the embedding layer, the embedding vector sequence $\mathbf{v} = [\mathbf{v}_1, \ldots, \mathbf{v}_T]$ is obtained. This is then fed into $L$ transformer blocks, and the calculation of the output of the $l$-th transformer block is defined as:
\begin{equation}
    \mathbf{h}_t^{(l)} = \text{Transformer-Block}^{(L)}(\mathbf{h}_t^{(l-1)}) \,,
\end{equation}
where $\text{Transformer-Block}$ denotes the transformer block in GPT2, $\mathbf{h}_t^{(l)}$ is the output of the $l$-th transformer block at time step $t$, and $h_t^{(0)} = \mathbf{v}_t$. The output of the L-th transformer block is then fed into a linear layer to predict the action at each time step.
The probability of each action at time step t is computed by the following equation:
\begin{equation}
    P(\tau_t|\tau_{0},\ldots ,\tau_{t-1}) = \text{Softmax}(W^e \mathbf{h}_t^{(l)} + \mathbf{b}^{out})\,,
\end{equation}
where $\tau_t$ is the combination of $r_t,s_t,a_t$ and then the prediction of action at time step t $a_t$ is splitting from $\tau_t$. $\text{Softmax}$ denotes the same $\text{Softmax}$ operation from the vanilla Transformer\cite{transformer}.

Similarly, like the standard language modeling objective, we can maximize the following likelihood which is conditioned on the past trajectory $\tau_{0},\ldots ,\tau_{t-1}$ and model parameters $\theta$:
\begin{equation}
    \label{eq:tau_likelihood}
    \mathcal{L} = \sum_{t=1}^T \log P(\tau_t|\tau_{0},\ldots ,\tau_{t-1};\theta)\,.
\end{equation}
Since the prediction of return and state is not necessary for this task, the action prediction is the only concern. Equation \ref{eq:tau_likelihood} is then modified as follows:   
\begin{equation}
    \label{eq:action_likelihood}
    \mathcal{L} = \sum_{t=1}^T \log P(a_t\mid r_0,s_0, a_0, r_1, s_1, a_1, \ldots,r_{t-1}, s_{t-1}, a_{t-1},r_{t},s_{t};\theta)\,.
\end{equation}
\subsection{Offline Pre-training}
Contrary to the approach of transformer encoder-based methods, which utilize a variety of proxy tasks for learning multi-modal representations, our methodology capitalizes on a sequence action prediction task for pre-training the model. The action label predicted by the model is rigorously compared against the ground truth action label from the dataset. This strategy is not merely aimed at replicating the expert behavior documented within the dataset; it also facilitates the understanding of the interdependencies between observations and actions within the sequence. This is achieved by harnessing the sequence modeling prowess of the GPT model.

\noindent
\textbf{Sequence Action Prediction (SAP).} The objective of the sequence action prediction task is to forecast the action at each time step, given the trajectory preceding the current state in the sequence. This aligns with the action prediction tasks commonly employed during the pre-training phase of encoder-based methodologies. A notable distinction in our approach is the elimination of the need for explicit history encoding, as the task is framed as a sequence decision-making problem (\cref{eq:action_likelihood}), with action predictions being made via a decoder-only transformer. Owing to this formulation and the implementation of a masked attention mechanism, the prediction of subsequent actions is intrinsically conditioned on the preceding trajectories.
We formulate this as a classification task featuring an action prediction head, maintaining the architectural design consistent with the HAMT model. Specifically, the action prediction head comprises two layers of the fully connected network.
Hence, we predict action probability for each navigable view in $O_t$ as the following equation:
\begin{equation}
\label{eq:action_prediction}
    p_t\left(s_{t,i}\right)=\frac{\exp \left(f_{\mathrm{SAP}}\left(s_{t,i}\right)\right)}{\sum_j \exp \left(f_{\mathrm{SAP}}\left(s_{t,j}\right)\right)}\,,
\end{equation} 
where $f_{\mathrm{SAP}}$ is the action prediction head, $s_{t,i}$ is the state defined in \cref{eq:state}.

Likewise, the objective is to minimize the negative log probability of the target view action
\begin{equation}
    s_*: L_{\mathrm{SAP}}=-\log p_t\left(s_*\right)\,,
\end{equation}
where $s_*$ denotes any state at timestep $t$ of ith view in the trajectory $p_t$ is the same definition in \cref{eq:action_prediction}.
\subsection{Online Fine-tuning}
Given that we distinguish between the objectives of exploration and exploitation by allocating them to the pre-training and fine-tuning phases, respectively, and that learning from expert demonstrations occurs during the pre-training phase, we promote exploration during the online fine-tuning phase through the entropy of the policy. Same as in online decision transformer\cite{zheng2022online}, the entropy of the policy is defined as:
\begin{equation}
\label{eq:ft_obj}
    \begin{gathered}
    H_\theta^{\mathcal{T}}[\mathbf{a} \mid \mathbf{s}, \mathbf{r}]=\frac{1}{K} \mathbb{E}_{(\mathbf{s}, \mathbf{r}) \sim \mathcal{T}}\left[H\left[\pi_\theta(\mathbf{a} \mid \mathbf{s}, \mathbf{r})\right]\right] \\
    =\frac{1}{T} \mathbb{E}_{(\mathbf{s}, \mathbf{r}) \sim \mathcal{T}}\left[\sum_{t=1}^T H\left[\pi_\theta\left(a_t \mid \mathbf{s}_{-T, t}, \mathbf{r}_{-T, t}\right)\right]\right]\,,
    \end{gathered}
\end{equation}
where $H\left[\pi_\theta\left(a_t\right)\right]$ denotes the Shannon entropy of the distribution $\pi_\theta\left(a_t\right)$. The policy entropy is related to the data
distribution $\mathcal{T}$, which is static in the offline pre-training phase but dynamic during fine-tuning as it depends on the online data acquired during exploration.
\section{Experiments}

\subsection{Datasets and Evaluation Metrics}
\textbf{Dataset.} We evaluate our model on Room-to-Room(R2R)\cite{r2r} dataset. R2R is a dataset which builds upon Matterport3D \cite{Matterport3D} and consists of s 90 photo-realistic houses with 10,567
panoramas. It contains 7,189 path trajectories; each trajectory is annotated with three instructions. The dataset is split into the train, val seen, and val unseen sets with trajectories in 61, 56, and 18 buildings, respectively. 
Buildings in the validation seen split are the same as the training split, while houses in the validation unseen split differ from the training split.

\noindent
\textbf{Evaluation metrics.} We use action prediction accuracy to monitor the pre-training phase. In the fine-tuning stage, we adopt standard metrics \cite{anderson2018evaluation} for the evaluation of the VLN task, which is the following:
\begin{enumerate}
    \label{sec:metrics}
    \item Trajectory Length (TL): the agent's navigated path in meters
    \item Navigation Error (NE): the average distance in meters between the agent's final position and the target
    \item Success Rate (SR): the percentage of trajectories that are successful, i.e., the agent stops within 3 meters of the goal location
    \item Success Rate normalized by the ratio between the length of the shortest path and the predicted path (SPL).
\end{enumerate}

\subsection{Implementation Details}
We retained the language transformer and vision transformer settings from HAMT\cite{hamt}. We adopt the GPT-2 base model\cite{gpt2} as the transformer decoder. We train the VLN-GPT  for 50k iterations using a learning rate of 5e-5 and a batch size of 64 on 1 NVIDIA RTX A6000 (5 hours) for the offline pre-training stage. In the online fine-tuning stage, The model is fine-tuned for 100k iterations with a learning rate of 1e-5 and batch size of
8 on the same NVIDIA RTX A6000 GPU. 
\subsection{Main Results}
\begin{table}[t!]
    \caption{Comparison of the online fine-tunig result with state-of-the-art methods on R2R dataset. The rows colored grey are transformer-encoder-based methods. The best results are marked in bold, with the second-best results underlined.}
    \label{tab: main result}
    \centering
    \begin{tabular}[!t]{@{}ccccccccc@{}}
    \toprule
    \multirow{2}{*}{Methods} & \multicolumn{4}{c}{Validation Seen} & \multicolumn{4}{l}{Validation Unseen} \\
     & TL & NE$\downarrow$ & SR$\uparrow$ & SPL$\uparrow$ & TL & NE$\downarrow$ & SR$\uparrow$ & SPL$\uparrow$ \\ \midrule
    Seq2Seq\cite{r2r} & 11.33 & 6.01 & 39 & - & 8.39 & 7.81 & 22 & - \\
    Babywalk\cite{zhu2020babywalk} & 10.36 & 5.10 & 54 & 50 & 10.75 & 6.43 & 39 & 34 \\
    PRESS\cite{li-etal-2019-robust} & 10.57 & 4.39 & 58 & 55 & 10.36 & 5.28 & 49 & 45 \\
    EnvDrop\cite{tan-etal-2019-learning} & 11.00 & 3.99 & 62 & 59 & 10.70 & 5.22 & 52 & 48 \\
    SF\cite{fried2018speaker} & - & 3.36 & 66 & - & - & 6.62 & 35 & - \\
    RelGraph\cite{relgraph} & 10.13 & 3.47 & 67 & 65 & 9.99 & 4.73 & 57 & 53 \\
    \rowcolor{gray!25} PREVALENT\cite{PREVALENT} & 10.32 & 3.67 & 69 & 65 & 10.19 & 4.71 & 58 & 53 \\ 
    \rowcolor{gray!25} RecBERT\cite{rnnbert}  & 11.13 & \underline{2.90} & \underline{72} & \underline{68} & 12.01 & \underline{3.93} & \underline{63} & \underline{57}   \\
    \midrule
    \rowcolor{green!10}\multicolumn{1}{l}{\ours} & 11.18 & \textbf{2.55} & \textbf{76} & \textbf{72} & 11.51 & \textbf{3.75} & \textbf{65} & \textbf{61} \\ \bottomrule
    \end{tabular}
\end{table}
To assess the efficacy of the pre-training phase, we benchmark the action prediction accuracy of the VLN-GPT model against state-of-the-art methods on the R2R dataset, specifically selecting those methods that incorporate the Sequence Action Prediction (SAP) task during pre-training. Accordingly, PREVALENT\cite{PREVALENT} and HAMT\cite{hamt} are chosen for this comparison. The findings, presented in Table \ref{tab: pre-train result}, indicate that the VLN-GPT model exhibits commendable performance in the pre-training phase, surpassing other leading methods in terms of action prediction accuracy. This underscores the value of integrating SAP tasks as supervisory signals in the pre-training phase for our decoder-only transformer model architecture.

\begin{table}[t!]
    \caption{Comparison of the pre-training result on action prediction accuracy of the R2R validation dataset. The best results are marked in bold.}
    \label{tab: pre-train result}
    \centering
    \begin{tabular}{@{}cclllclll@{}}
        \toprule
        \multirow{2}{*}{Methods} & \multicolumn{4}{c}{Validation Seen} & \multicolumn{4}{l}{Validation Unseen} \\
         & \multicolumn{4}{c}{SAP$\uparrow$} & \multicolumn{4}{c}{SAP$\uparrow$} \\ \midrule
        PREVALENT \cite{PREVALENT} & \multicolumn{4}{c}{62} & \multicolumn{4}{c}{58} \\
        HAMT \cite{hamt} & \multicolumn{4}{c}{70} & \multicolumn{4}{c}{68} \\ \midrule
        \rowcolor{green!10}\multicolumn{1}{l}{\ours} & \multicolumn{4}{c}{\textbf{78}} & \multicolumn{4}{c}{\textbf{72}} \\ \bottomrule
    \end{tabular}
\end{table}

Subsequently, we undertake experiments to evaluate the online fine-tuning performance using the metrics outlined in \cref{sec:metrics} on the R2R dataset. Table \ref{tab: main result} juxtaposes the VLN-GPT model's performance against previous VLN methodologies within the R2R benchmark. The outcomes reveal that, despite a simplified structure and training approach, our model achieves promising results comparable to other transformer-encoder-based methods in SR across both the validation seen dataset and SPL metrics for both the validation seen and unseen datasets.
\subsection{Ablation Studies}
In this section, we undertake a series of experiments to assess the influence of various components within our VLN-GPT model on VLN task performance. Our evaluation begins with an examination of the sequential modeling setting's effectiveness. Subsequently, we explore the impact of the transformer block count on navigation outcomes, thereby assessing how changes in parameter scale might influence results. We also delve into the contributions of both the pre-training and fine-tuning phases to the overall success of the VLN task. Additionally, the effects of model weight initialization and the transformer decoder architecture on performance are investigated.
Due to space limitations, ablation studies on the impact of model weights initialization and transformer decoder architecture are deferred to the Appendix.

\noindent
\textbf{Impact of sequential modeling.} To assess the model's performance under non-sequential settings, we adjusted the history trajectory length in \cref{eq:sequential_policy} to 1, thereby conditioning the action prediction on single-step observations without reliance on trajectory history. The outcomes of this modification are presented in Table \ref{tab:seq_ablation}. The findings illustrate that sequential modeling significantly enhances the performance across all evaluation metrics for both the validation seen and unseen datasets. Specifically, the SR and SPL both exhibit a 7\% increase in the validation seen dataset compared to the non-sequential approach. In the case of the validation unseen dataset, the improvements are even more pronounced, with SR and SPL experiencing relative increases of 7\% and 8\%, respectively. This analysis underscores the superiority of sequential modeling over non-sequential approaches in enhancing task performance.
\begin{table}[t!]
    \caption{Ablation Studies on Sequential Modeling: For the non-sequential setting, the sequence length is adjusted to 1, whereas in the sequential setting, we maintain a sequence length of $T$ as outlined in \cref{eq:ft_obj}. The evaluations are conducted using the R2R validation dataset, with the optimal outcomes highlighted in bold.}
    \label{tab:seq_ablation}
    \centering
    \begin{tabular}{@{}ccccc@{}}
    \toprule
    \multirow{2}{*}{Sequential modeling} & \multicolumn{2}{c}{Validation Seen} & \multicolumn{2}{c}{Validation Unseen} \\
     & SR & SPL & SR & SPL \\ \midrule
     $\times$ & 69.53 & 65.87 & 58.65 & 53.42 \\
     \rowcolor{green!10}\checkmark  & \textbf{76.48} & \textbf{72.18} & \textbf{65.14} & \textbf{61.09} \\ \bottomrule
     \vspace{-0.8cm}
    \end{tabular}
\end{table}

\begin{figure}[!t]
    \vspace{-0.40cm}
    \centering
    \begin{subfigure}{0.49\textwidth}
        \centering
        \includegraphics[width=\textwidth]{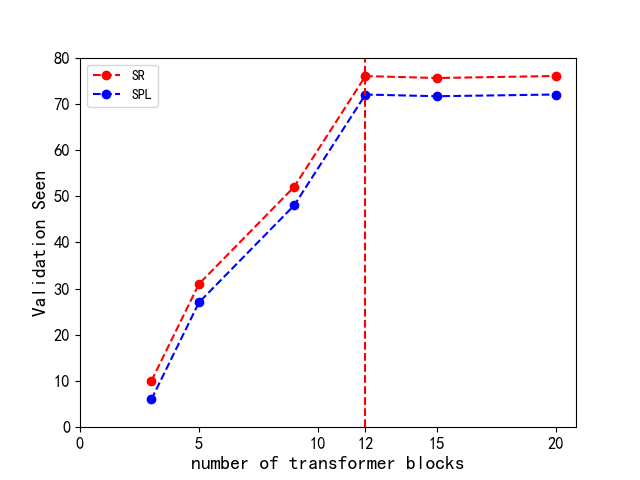}
        \caption{SR and SPL metrics derived from models with varying numbers of transformer blocks on the R2R validation seen dataset are illustrated. The horizontal axis denotes the number of transformer blocks, while the vertical axis quantifies the SR and SPL metrics.}
        \label{fig:sub1}
    \end{subfigure}
    \hfill
    \begin{subfigure}{0.49\textwidth}
        \centering
        \includegraphics[width=\textwidth]{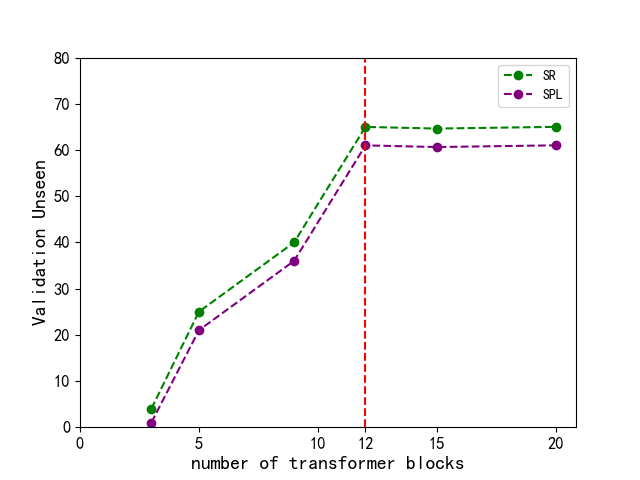}
        \caption{SR and SPL metrics derived from models with varying numbers of transformer blocks on the R2R validation unseen dataset are illustrated. The horizontal axis denotes the number of transformer blocks, while the vertical axis quantifies the SR and SPL metrics.}
        \label{fig:sub2}
    \end{subfigure}
    \caption{Success Rate (SR) and Success weighted by Path Length (SPL) outcomes from GPT models with varying parameter scales on the R2R validation dataset are depicted in \cref{fig:sub1} for SR and \cref{fig:sub2} for SPL, respectively. To create parameter scale variants of the GPT model, we adjust the number of transformer blocks. However, due to computational power constraints, experiments involving more than 20 transformer blocks are unfeasible.}
    \label{fig:parameter_scale}
    \vspace{-0.8cm}
\end{figure}

\noindent
\textbf{Impact of model parameter scale.} We explore the effect of the model parameter scale on Vision-and-Language Navigation (VLN) task performance by varying the number of transformer blocks within the model. Our experimental framework encompasses GPT models equipped with 3, 5, 9, 12, 15, and 20 layers of transformer blocks, with 12 representing the standard base model configuration for GPT2 and 20 constituting the medium setup. Computational resource constraints precluded the investigation of models with a higher number of transformer blocks.

The findings, as detailed in \cref{fig:parameter_scale}, indicate a consistent improvement in both Success Rate (SR) and Success weighted by Path Length (SPL) up to the 12-transformer-block threshold. Beyond this point, the SR and SPL metrics exhibit a plateau, hovering around 70\% for models exceeding the parameter scale of the base GPT configuration. This observation suggests that the GPT base model configuration suffices for VLN tasks on the R2R dataset. Nonetheless, the potential for enhanced performance in larger datasets remains, where models with increased parameter counts may yield superior results.


\begin{wraptable}{r}{0.55\textwidth}
\vspace{-0.2cm}
\caption{Ablation Studies on Pre-training and Fine-tuning: "PT" and "FT" denote pre-training and fine-tuning, respectively. The outcomes derived from these processes are evaluated using the R2R dataset. The best results are marked in bold.}
    \label{tab:pt_ft_ablation}
    \centering
    \begin{tabular}{@{}cccccc@{}}
    \toprule
    \multirow{2}{*}{PT} & \multirow{2}{*}{FT} & \multicolumn{2}{c}{Validation Seen} & \multicolumn{2}{c}{Validation Unseen} \\
     &  & SR $\uparrow$ & SPL$\uparrow$ & SR$\uparrow$ & SPL$\uparrow$ \\ \midrule
     \checkmark & $\times$ & 64.08 & 62.17 & 52.34 & 49.67 \\
     $\times$ & \checkmark & 70.12 & 65.58 & 64.76 & 58.84 \\
    \rowcolor{green!10} \checkmark & \checkmark & \textbf{76.48} & \textbf{72.18} & \textbf{65.14} & \textbf{61.09} \\ \bottomrule
    \vspace{-6mm}
    \end{tabular}
\end{wraptable}
\noindent
\textbf{Impact of pre-training and fine-tuning.} Table \ref{tab:pt_ft_ablation} displays outcomes across various training phases. The initial row details the model's performance following exclusively pre-training, while the subsequent row delineates the performance after direct online fine-tuning. Observations indicate inferior performance from sole pre-training compared to exclusive fine-tuning. This discrepancy stems from the pre-training phase's reliance on a static offline trajectory dataset, which hinders the model's ability to assimilate dynamically acquired online data during exploration. The penultimate row illustrates the model's performance when combining both pre-training and fine-tuning, marking the apex of achievement. Such results validate the integral efficacy of incorporating both pre-training and fine-tuning stages within the VLN-GPT model's developmental process.
\section{Conclusion}
This paper introduces the pioneering decoder-only transformer architecture and a sequential approach to vision-and-language navigation, coined as the Vision-and-Language Navigation Generative Pretrained Transformer (VLN-GPT). Our technique adeptly captures the interrelations among returns, states, and actions within trajectories and facilitates multimodal action predictions through a sequential decision-making process, thereby obviating the need for a history encoder found in other transformer-based studies. We have conceptualized a novel pre-training and fine-tuning framework tailored to the VLN task, which distinctly delineates the objectives of exploration and exploitation into offline pre-training and online fine-tuning phases, consequently simplifying the model's training complexity. Our approach is validated and demonstrates encouraging outcomes relative to the SOTA transformer encoder architectures.

\noindent
\textbf{Limitation.} 
The relatively modest scale of datasets within the VLN domain, as opposed to the expansive datasets common in Natural Language Processing (NLP), constrains the efficacy of larger transformer models characterized by elevated parameter counts, thereby impeding enhanced performance. In the future, the potential of VLN-GPT to navigate longer trajectories and process increasingly complex instructions warrants further exploration. Future research will delve into the advantages of pre-training on more extensive navigation datasets, alongside assessing the impact of models with larger parameter scales when applied to broader datasets.

\newpage
\appendix
In this appendix, section \ref{sec:exp} provides ablation studies on weights initialization and transformer decoder architecture, and section \ref{sec:qualitative results} illustrates qualitative results.

\section{Experiments}\label{sec:exp}
\noindent
\textbf{Impact of model weights initialization.} 
We delve deeper into the role of weight initialization, examining if pre-trained weights from language data enhance VLN task performance. The content of our experiment is to compare one model initialized with pre-trained weights from language data available on Hugging Face \cite{gpt2model} against another using the default initialization method outlined in the Hugging Face Transformers \cite{huggingface2023gpt2} GPT2 framework. Evaluation of these models on the R2R validation seen and unseen datasets (seen in \cref{tab:weights_init}) reveals nuanced effects: pre-trained language weights marginally improve performance on unseen data but slightly detract from seen data outcomes. Specifically, the improvement in SR and SPL for unseen data does not exceed 0.5\%, whereas the reduction in these metrics for seen data is confined to a maximum of 0.5\%. The disparity between the outcomes of the two weight initialization strategies is minimal, under 0.5\%, leading to the conclusion that language data-derived weights offer minimal contribution to multimodal decision-making in the VLN task. As models converge upon the data, the end results of models initialized via different methods are expected to be comparably effective.

\noindent
\textbf{Impact of transformer decoder architecture.} 
We embark on a series of experiments to explore the influence of various transformer decoder architectures on the outcomes of the Vision-and-Language Navigation (VLN) task. Specifically, we examine the performance of OPT \cite{opt}, Llama \cite{llama}, and GPT2 as different transformer decoder variants. To ensure a fair comparison, the base model versions of OPT and GPT2 are selected, each equipped with 12 layers of transformer blocks and 12 heads per block, encapsulating approximately 125M parameters. Correspondingly, we modify the Llama model to align with this parameter scale by reducing its transformer blocks and heads to 12, thereby standardizing the parameter count across Llama, GPT, and OPT models.
Analysis of the results presented in \cref{tab:decoder_types} reveals that, despite variances in decoder architecture, all models exhibit nearly identical SR and SPL across both validation seen and unseen datasets, with a negligible discrepancy not exceeding 0.05. Notwithstanding these similarities, the GPT model marginally surpasses its counterparts in performance. A plausible explanation for the observed phenomenon is that the disparities among these models are constrained by the relatively limited volume of VLN data available. Consequently, informed by these findings, we select the GPT architecture as the cornerstone for our VLN agent model.

\begin{table}[!t]
\caption{Ablation Studies on Model Weight Initialization: For this analysis, we utilize the pre-trained weights of the GPT2 base model on language data from Hugging Face \cite{gpt2model}, and compare it with the default weight initialization method found in the Hugging Face Transformers \cite{huggingface2023gpt2} GPT2 implementation. The comparison of these two initialization methods is conducted on the R2R validation dataset, covering both the validation seen and unseen datasets. The best results are marked in bold.}
\centering
\label{tab:weights_init}
\begin{tabular}{@{}ccccc@{}}
\toprule
\multirow{2}{*}{Language pre-trained weights} & \multicolumn{2}{c}{Validation Seen} & \multicolumn{2}{c}{Validation Unseen} \\
 & SR & SPL & SR & SPL  \\ \midrule
\checkmark & 75.84 & 71.96 & \textbf{65.33} & \textbf{61.27}\\
\rowcolor{green!10} $\times$ & \textbf{76.23} & \textbf{72.18} & 65.09 & 61.16\\ \bottomrule
\end{tabular}
\end{table}

\begin{table}[!t]
\caption{Ablation Studies on Transformer Decoder Architecture: To ensure equitable comparisons, we utilize the base models of GPT and OPT, each comprising 12 layers of Transformer blocks. To align the Llama model with the same parameter scale, we adjust the number of Transformer blocks accordingly; herein, "Llama*" denotes this adjusted version of the Llama model. This methodology facilitates a consistent comparison across different models on the R2R dataset, under identical conditions. Optimal outcomes are highlighted in bold for clarity.}
\centering
\label{tab:decoder_types}
\begin{tabular}{ccccc}
\toprule
\multirow{2}{*}{Model Architecture} & \multicolumn{2}{c}{Validation Seen} & \multicolumn{2}{c}{Validation Unseen} \\
 & SR & SPL & SR & SPL \\ \midrule
OPT & 76.40 & 72.11 & 65.17 & 61.21 \\
Llama* & 76.37 & 72.09 & 65.19 & 61.19 \\
\rowcolor{green!10} GPT & \textbf{76.42} & \textbf{72.15} & \textbf{65.23} & \textbf{61.22} \\ \bottomrule
\end{tabular}
\end{table}

\section{Qualilitive Results}\label{sec:qualitative results}
\begin{figure}[t]
  \centering
  \begin{subfigure}[b]{0.45\textwidth} 
    \includegraphics[width=\textwidth]{example_v2.pdf}
    \caption{Example 1 from R2R dataset}
    \label{fig:eg1}
  \end{subfigure}
  \hfill 
  \begin{subfigure}[b]{0.45\textwidth} 
    \includegraphics[width=\textwidth]{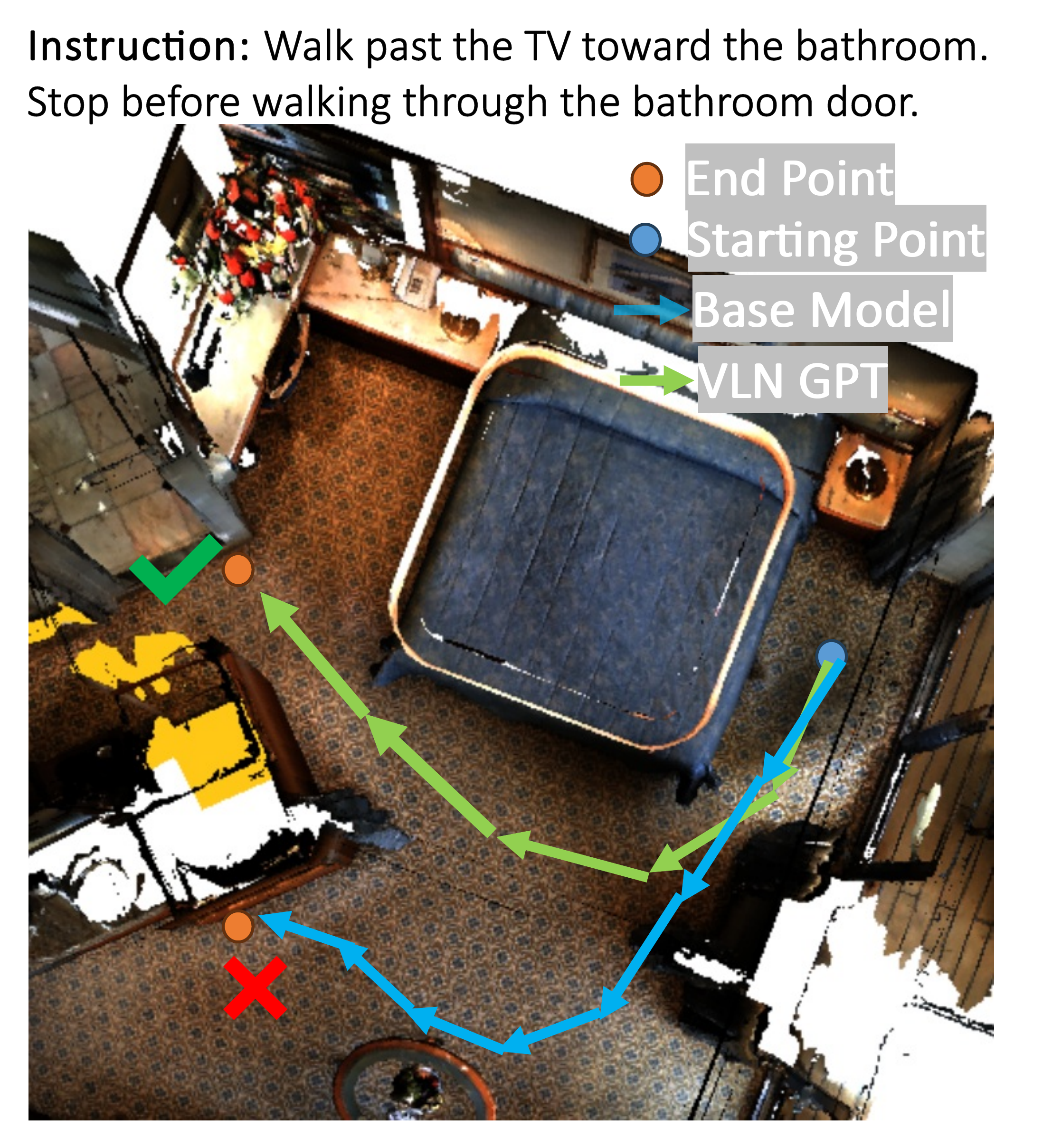}
    \caption{Example 2 from R2R dataset}
    \label{fig:eg2}
  \end{subfigure}

  \begin{subfigure}[b]{0.9\textwidth} 
    \centering
    \includegraphics[width=\textwidth]{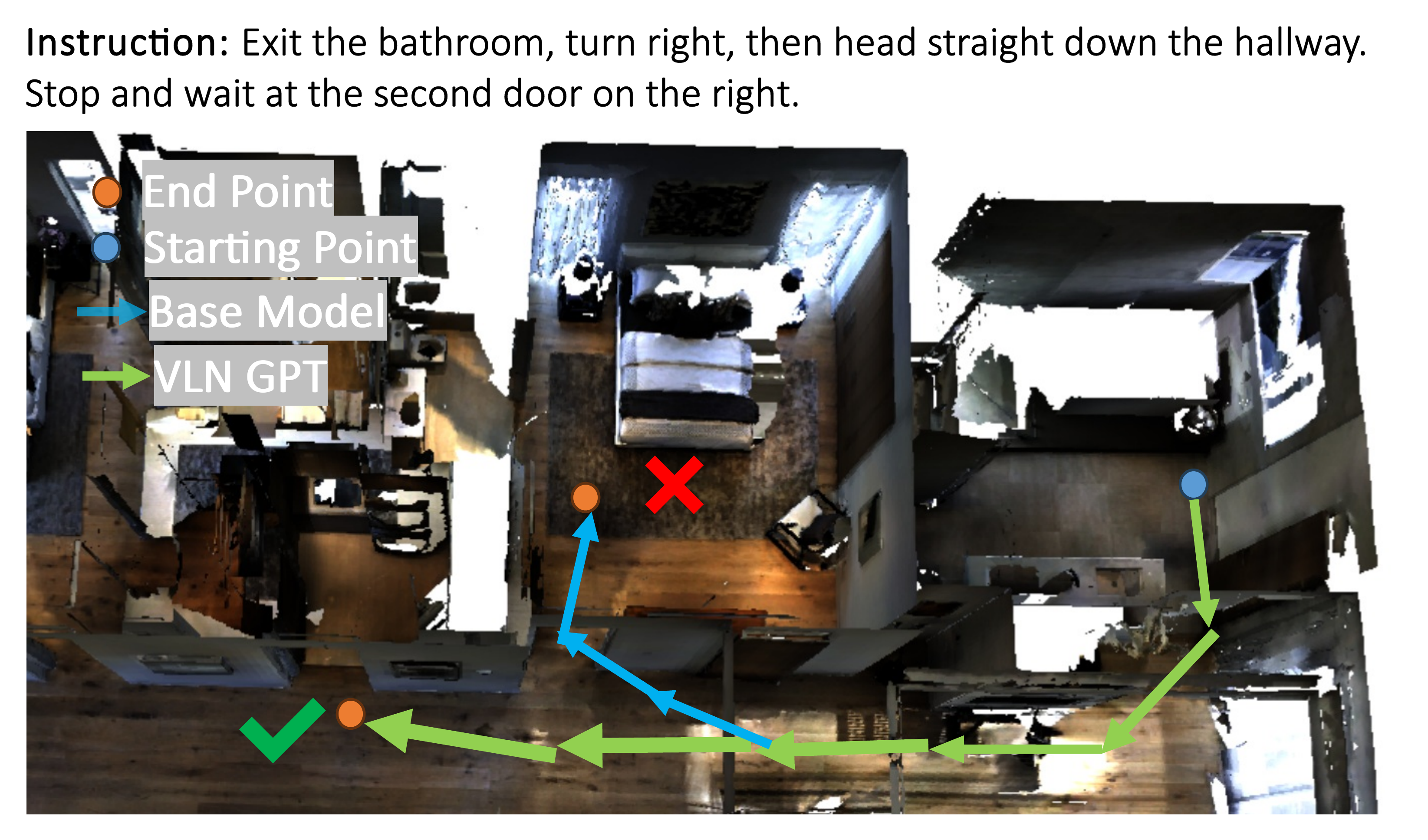}
    \caption{Example 3 from R2R dataset}
    \label{fig:eg3}
  \end{subfigure}

  \caption{Demonstration of examples from the R2R validation dataset. The sentence at the top is the instruction of this example. The background image is an overhead view of the navigation room. The green arrows denote the trajectory of our VLN-GPT agent, and the blue one is the trajectory from PREVALENT as the base model with transformer encoder architecture. The blue point in the figure is the starting point of the trajectory, and the orange point is the endpoint. The text label \textcolor{green}{$\checkmark$} means the agent successfully reaches the intended target through the trajectory, and the text label \textcolor{red}{$\times$} means the agent fails to navigate to the target.}
  \label{fig:qualitative_example}
\end{figure}

We illustrate the qualitative outcomes by selecting a specific set of instructions from the R2R validation dataset. Figure \ref{fig:qualitative_example} presents the navigation trajectory as predicted by our VLN-GPT model compared to the base model from PREVALENT. Notably, our VLN-GPT agent adeptly navigates in accordance with the instructions to reach the designated destination, in contrast to the base model agent, which does not accomplish this task. This comparison highlights the superior ability of our VLN-GPT approach to effectively capture the multi-modal relationships and complexities inherent in the instructions and trajectory.


\bibliographystyle{unsrt}  
\bibliography{references}

\end{document}